\documentclass[sigconf]{acmart}

\usepackage[abbreviations]{glossaries-extra}
\usepackage{subcaption}

\newabbreviation{cppn}{CPPN}{Compositional Pattern Producing Network}
\newabbreviation{neat}{NEAT}{Neuroevolution of Augmenting Topologies}
\newabbreviation{er}{ER}{evolutionary robotics}
\newabbreviation{cpg}{CPG}{Central Pattern Generator}
\newabbreviation{ai}{AI}{artificial intelligence}
\newabbreviation{ann}{ANN}{artificial neural network}
\newabbreviation{ea}{EA}{evolutionary algorithm}
\newabbreviation{ec}{EC}{embodied cognition}
\newabbreviation{la}{LA}{linear actuator}

\AtBeginDocument{%
  \providecommand\BibTeX{{%
    \normalfont B\kern-0.5em{\scshape i\kern-0.25em b}\kern-0.8em\TeX}}}

\setcopyright{acmcopyright}
\copyrightyear{2022}
\acmYear{2022}
\acmDOI{10.1145/nnnnnnn.nnnnnnn}

\acmConference[]{}{}{}
\acmPrice{}
\acmISBN{}

\renewcommand\footnotetextcopyrightpermission[1]{} 

\begin{document}

\title{The Effects of the Environment and Linear Actuators on Robot Morphologies}

\author{Steven Oud}
\authornote{Both authors contributed equally to this research.}
\affiliation{%
  \institution{Vrije Universiteit Amsterdam}
  \city{Amsterdam}
  \state{Noord-Holland}
  \country{The Netherlands}
  }
\email{s.oud2@student.vu.nl}

\author{Koen van der Pool}
\affiliation{%
  \institution{Vrije Universiteit Amsterdam}
  \city{Amsterdam}
  \state{Noord-Holland}
  \country{The Netherlands}
  }
\email{k.vander.pool@student.vu.nl}

\renewcommand{\shortauthors}{Oud and van der Pool}

\begin{abstract}
The field of \acrlong{er} uses principles of natural evolution to design robots.
In this paper, we study the effect of adding a new module inspired by the skeletal muscle to the existing RoboGen framework: the linear actuator.
Additionally, we investigate how robots evolved in a plain environment differ from robots evolved in a rough environment.
We consider the task of directed locomotion for comparing evolved robot morphologies.
The results show that the addition of the linear actuator does not have a significant impact on the performance and morphologies of robots evolved in a plain environment.
However, we find significant differences in the morphologies of robots evolved in a plain environment and robots evolved in a rough environment.
We find that more complex behavior and morphologies emerge when we change the terrain of the environment.
\end{abstract}



\maketitle

\pagestyle{plain}

\section{Introduction}
Nature has long been an inspiration for scientists and engineers to design systems, machines, and robots to solve complex problems. Already in the 1480s, Leonardo da Vinci designed the ornithopter, a flying machine inspired by the flapping of a bird to make humans fly. Despite never being made, the ornithopter is an example of a machine whose design is based on natural elements.

\Gls{er}, an embodied approach to artificial intelligence, is a branch of research whose foundations are derived from nature. This field is inspired by the process that underpins life on earth: evolution. The aim is to utilize the principles of natural evolution, such as selection, variation, and heredity, in an \gls{ea} to evolve the morphology and controllers of robots \cite{doncieux2015evolutionary}. By doing so, robots with high fitness are obtained, i.e. robots that show high-quality behavior for a given task measured by a predefined metric \cite{eiben2003introduction, doncieux2015evolutionary}. The ultimate long-term vision of \gls{er} is to automate the design, construction, and distribution of autonomous robots \cite{bongard2013evolutionary, doncieux2015evolutionary}.

An important theory behind the deployment of \glspl{ea} is embodied cognition, which states that the interaction between the body, brain, and environment is the source of intelligence \cite{bongard2013evolutionary,pfeifer2006body, anderson2003embodied, beer2008dynamics, weigmann2012does}. This theory can be applied to \gls{er}, where a robot's fitness is affected by the interaction between the controllers, morphology, and environment \cite{bongard2013evolutionary, miras2020environmental, weigmann2012does}. Therefore, it can be argued that it is of importance to investigate all three of these elements. Despite this, most of the research has focused on the evolution of the controllers and has given little attention to the morphology and environment \cite{auerbach2014environmental, miras2020environmental, pfeifer2009morphological}.

This research focuses on the morphology of robots that are evolved for the task of directed locomotion. The initial robots are built out of four modules from the RoboGen framework \cite{auerbach2014robogen}. In order to expand the morphological search space, we add an extra module to the current framework. This is done to investigate whether the robot's task performance, namely locomotion, improves if it has more options to evolve its morphology. Besides, we are interested if the morphologies with the extra module are significantly different from those that do not include the extra module. When designing this module, we were inspired by nature, just as da Vinci. Humans and animals are known to locomote successfully in diverse and unstructured environments. An important element behind this is the skeletal muscle, as it outputs the forces necessary for locomotion \cite{king2004force, lieber1999skeletal}. The properties of the muscle inspired the design of multiple actuators that generate force, which is a function of velocity, length, and activation level, in linear motion \cite{hannaford2001bio, klute1999mckibben}. Therefore, we introduce the new module to the robot's morphology based on the skeletal muscle: the linear actuator. This eventually led to the first research question addressed in this study:

\begin{itemize}
  \item [$\bullet$] What is the effect of adding the linear actuator module on the performance and morphology of the robots?
\end{itemize}

Besides investigating the effects of an extra module on the fitness and morphology of a robot, we also want to focus on the third element: the environment. On its own, little can be said about the environment. It is more interesting to investigate the influence of the environment on the evolution of the controllers and the morphology. In previous research, little attention has been paid to the influence of the environment on the robot’s morphology and therefore it is not fully understood \cite{auerbach2014environmental, miras2020environmental, pfeifer2009morphological}. Here, we will explore this by evolving the robots, with the initial RoboGen framework of four modules, in two different environments. The first environment has a surface that is equal to the aforementioned plain surface. The second environment is considered more challenging and has a rough surface built in Blender, an open-source software toolkit for 3D graphics~\cite{blender}. We evolve robots in both environments in order to answer the study's second research question:

\begin{itemize}
  \item [$\bullet$]How do the morphologies of robots evolved in a plain environment differ from the morphologies of robots evolved in a rough environment?
\end{itemize}

This paper is divided into six sections. The second section provides a review of all the important work on the linear actuator in ER research, as well as the effect of the environment on robot morphologies. The robot framework, including the morphology, controller, representation, and morphological descriptors, is the focus of the third section. Section four describes the experimental setup, whereas section five covers the experimental results and discussion. The conclusion and suggestions for future work are discussed in the final section.

\section{Related Work}
In our research, a linear actuator module is added to the robot's morphology. Research in \gls{er}, particularly done in hardware, has also introduced a linear actuator to the morphology of its robot. In \cite{nygaard2018real}, it is examined how a four-legged robot adapts itself for the task of locomotion when subjected to various hardware constraints by evolving both its controller and morphology. The legs are partially made of linear actuators, allowing them to expand. The results show that the robot can successfully adapt itself when faced with limitations in the hardware. In a follow-up study \cite{nygaard2021environmental}, the four-legged robot is evolved for the task of locomotion over different surfaces. It was observed that the robot is successful in evolving its controller and morphology in order to locomote over these different terrains. In an additional study, Nyagaard et al. \cite{nygaard2019self} confirmed these results and found that ``mechanically self-modifying robots may perform better in dynamic environments by adapting morphology as well as control to new conditions." Although the design of the robot is the result of human work, these results are promising for adding a linear actuator to the modules of the robot's morphology.

In the field of \gls{er}, robots are often evolved in an environment for the task of locomotion where the focus is on the evolution of the robot's controller \cite{auerbach2014environmental, miras2020environmental, pfeifer2009morphological}. Karl Sims \cite{sims1994evolving} introduced the approach to evolve both the controller and morphology of virtual creatures simultaneously. While the evolution of the creatures' morphologies and controllers was studied, the influence of the environment on this process was not investigated. Other studies in \gls{er} (e.g. \cite{lipson2000automatic, hornby2001body, auerbach2010dynamic, auerbach2011evolving, chee2014simultaneous}) followed Sims' approach to evolve the controller and morphology at the same time. However, \cite{auerbach2012relationship2} pointed out that the most widely studied task in this area of research has been displacement velocity over a flat terrain. While examining this task yielded fascinating results, Auerbach et al. \cite{auerbach2012relationship2} argued that "it suffers from its simplicity". A simple morphology, consisting of only a few modules, turns out to be sufficient for an autonomous robot to succeed in such a task environment. 

Therefore, Auerbach et al. \cite{auerbach2012relationship2} evolved robots in more complex environments to investigate whether the complexity of the morphology increases. The findings indicate that evolving robots in a more complex environment leads to the evolution of robots with more complex morphologies. The same results were also found in \cite{auerbach2014environmental}, where a cost was placed on morphological complexity. In another study, Auerbach et al. \cite{auerbach2012relationship} examined whether the mechanical complexity of evolved robots scales with the complexity of the task environment, where they define mechanical complexity as ``the number of mechanical degrees of freedom in an evolved robot". In their results, they observed that, while the morphological complexity increased, robots evolved in more complex environments have a lower mechanical complexity than robots evolved in a simpler environment. Auerbach et al. \cite{auerbach2012relationship} state that this finding is counter-intuitive, as the various forms of morphological complexity were assumed to be correlated, but are presumably orthogonal. However, by expressing the morphological complexity in one value, it is not possible to investigate how different properties of the environment impact the robot's morphology. As a result, it is unable to explain how distinct morphologies evolved from environments with similar complexity.

In order to fill this gap, Miras et al. \cite{miras2018search} introduced a framework of eight morphological descriptors to quantify the differences in the morphologies. This framework is also used in our research. In \cite{miras2019effects}, the framework was used to investigate whether evolving autonomous robots in distinct environments for the task of locomotion leads to morphological changes. The robots were evolved in three different environments, namely plain, obstacles, and titled. The robots that evolved in the titled environment had significantly different morphologies from those that evolved in the other two environments. However, the same `snake'-like morphologies arose in the plain and obstacles environments, suggesting that different environments can lead to the same morphologies. These results are surprising since they contradict the main principles of evolution. Further investigation revealed that the `snake'-like morphologies were the result of a bias in the reproduction operators of the used genetic encoding, namely the L-System \cite{10.3389/frobt.2021.672379}.

Van der Pool et al. used the aforementioned framework to investigate the effects of environmental differences on the morphologies of robots evolved for the task of locomotion \cite{pool_eiben_2022}. In particular, they were interested in the morphological differences of robots evolved in an Earth-like environment to those evolved under Moon-like conditions. The Earth-like benchmark scenario has a flat terrain with the surface gravity level of the Earth. To produce a Moon-like environment, two environmental properties are changed: the surface gravity level is reduced, and the flat terrain is replaced with a NASA-created 3D representation of the Apollo 14 moon landing site. The results show that changing only one of the environmental properties does not lead to morphological changes. However, when both properties are changed, it turns out that robots evolved in the Moon-like environment are usually bigger in size, have fewer limbs, and have a less space-filling shape than those evolved in Earth-like conditions. A result that is in line with one of the principles of evolution, namely that evolution in different environments leads to different morphologies.

\section{Robot Framework}
\subsection{Morphology} \label{sec:robot-morphology}
\begin{figure}[h]
  \centering
  \includegraphics[width=\linewidth]{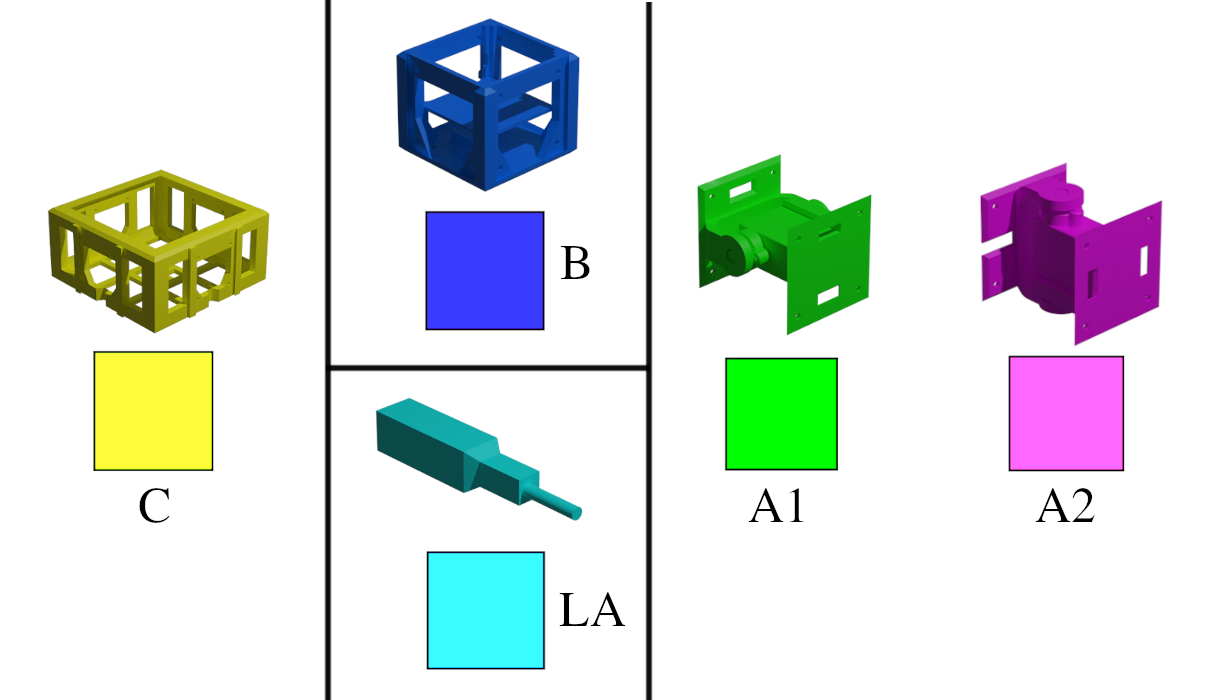}
  \caption{
  Robot modules: core component (C) holds the controller board; fixed brick (B) is the main structural component; active hinges (A1 and A2) have servo motor joints on horizontal and vertical axes respectively; \gls{la} produces straight push and pull movements.
  C and B have attachment slots on their four lateral faces; A1, A2, and LA have attachment slots on their two opposite lateral faces.
  }
  \label{fig:robot-modules}
\end{figure}

The phenotypes of the robot's morphologies consist of modules based on RoboGen \cite{auerbach2014robogen} (Figure~\ref{fig:robot-modules}).
This work expands on RoboGen by adding a linear actuator module.
Robot morphologies consist of one core component and several other components with a maximum of ten total components (including the core component).
Any component can be attached to any other component's attachment slot.
The fixed brick component can additionally be rotated by $90^\circ$, allowing the morphology to extend vertically.

\subsection{Controller}
The phenotype of the robot's controller is a hybrid \gls{ann} called recurrent central pattern generator perceptron~\cite{miras2019effects}.
It is based on \glspl{cpg} --- which are biological neural circuits that can produce rhythmic patterns without requiring rhythmic inputs~\cite{bucher2015central}.
\gls{cpg}-based controllers have been shown to successfully produce locomotion in animals and (modular) robots~\cite{ijspeert2008central,lan2018directed}.

\subsection{Representation and Operators}
For both the representation of the morphology and controller of the robot we use \glspl{cppn} as genotypes.
The morphology of a robot is built by iterating over every coordinate in the robot's morphology grid, for with the \gls{cppn} outputs the type of module to put in that coordinate.
Likewise, for the controller, we iterate over all pairs of connected \gls{cpg} nodes, for which the \gls{cppn} will output the corresponding weight.
The \glspl{cppn} are evolved using \gls{neat}, and for mutation and crossover, the standard \gls{neat} operators as described in \cite{stanley2002evolving} are used. 

\subsection{Morphological Descriptors}
In order to quantify and compare morphological properties, the following eight morphological descriptors proposed in \cite{miras2018search} are used:
\begin{enumerate}
    \item \textbf{Branching:} Ratio of the number of fully connected modules to the maximum possible number of modules that can be attached on four faces.
    This descriptor measures how spread out the components of the body are.
    \item \textbf{Coverage:} Describes how well the rectangular envelope around the morphology is filled.
    \item \textbf{Relative number of joints:} Ratio of the number of joints to the maximum possible number of joints in the morphology.
    \item \textbf{Relative number of limbs:} Ratio of the number of modules that have only one face attached to another module (except for the core component) to the maximum possible number of modules with only one face attached in the morphology.
    \item \textbf{Relative length of limbs:} Length of the limbs in the morphology.
    \item \textbf{Proportion:} The 2D ratio of the body.
    This is given by dividing the shortest side of the morphology by its longest side.
    \item \textbf{Absolute size:} Number of modules in the morphology.
    \item \textbf{Symmetry:} The reflexive symmetry of the body with the core component as the referential center.
\end{enumerate}

\section{Experimental Setup}
The experiments in this report were conducted through simulation.
Simulations are run using the Revolve framework\footnote{Our adjusted version of Revolve used for these experiments is available at \url{https://github.com/soudy/revolve/tree/roboreptiles}}~\cite{hupkes2018revolve}, using Gazebo~\cite{1389727} as physics simulator.
We carried out two types of experiments.
First, we investigate the effect that adding the linear actuator module has on the robot morphologies and fitness, and second, we compare the morphologies of the robots evolved in a plain and rough environment.
The second experiment extends the work done in \cite{miras2019effects} by experimenting with another type of environment, a different robot design space that can produce 3D morphologies through rotating modules (as described in Section \ref{sec:robot-morphology}), and a different representation.
In both experiments, the same task is considered: directed locomotion.

\subsection{Environments}
\begin{figure}[ht]
    \centering
	\begin{subfigure}{1\linewidth}
    	\centering
    	\includegraphics[width=1\linewidth]{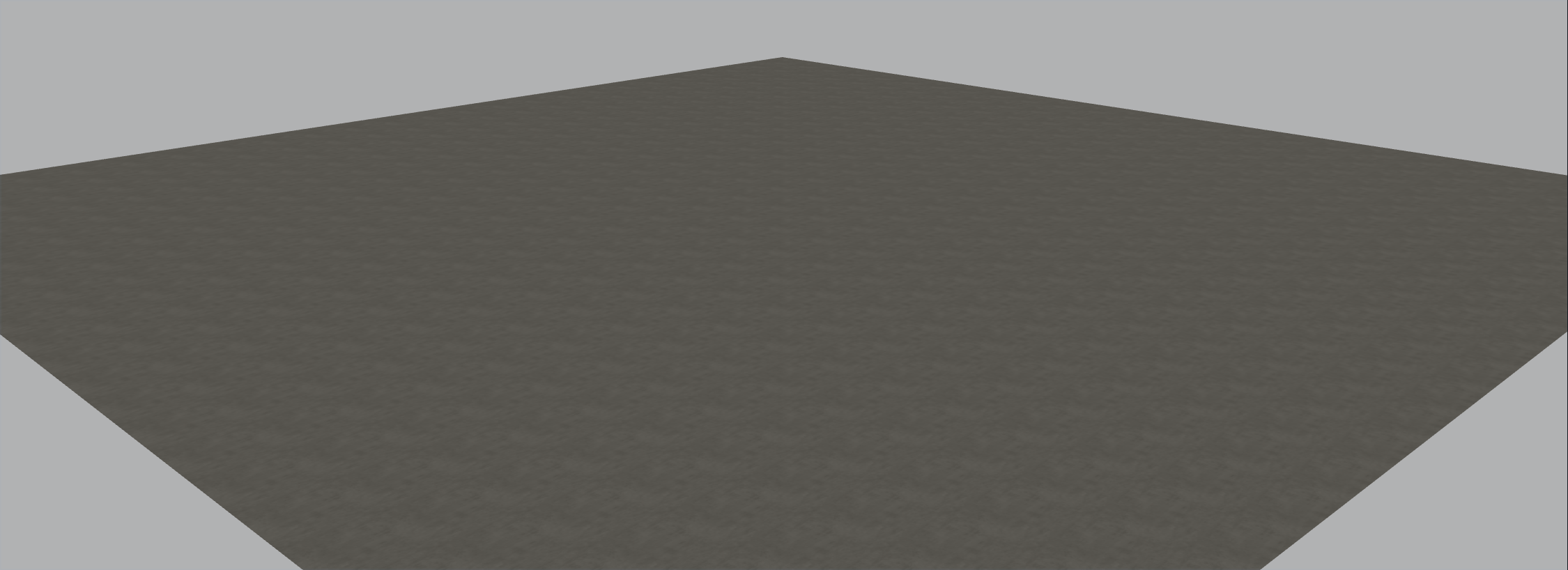}
    	\caption{
    	    Plain environment
        }
        \label{fig:plane-env}
    \end{subfigure}
	\hfill
	\begin{subfigure}{1\linewidth}
    	\centering
    	\includegraphics[width=1\linewidth]{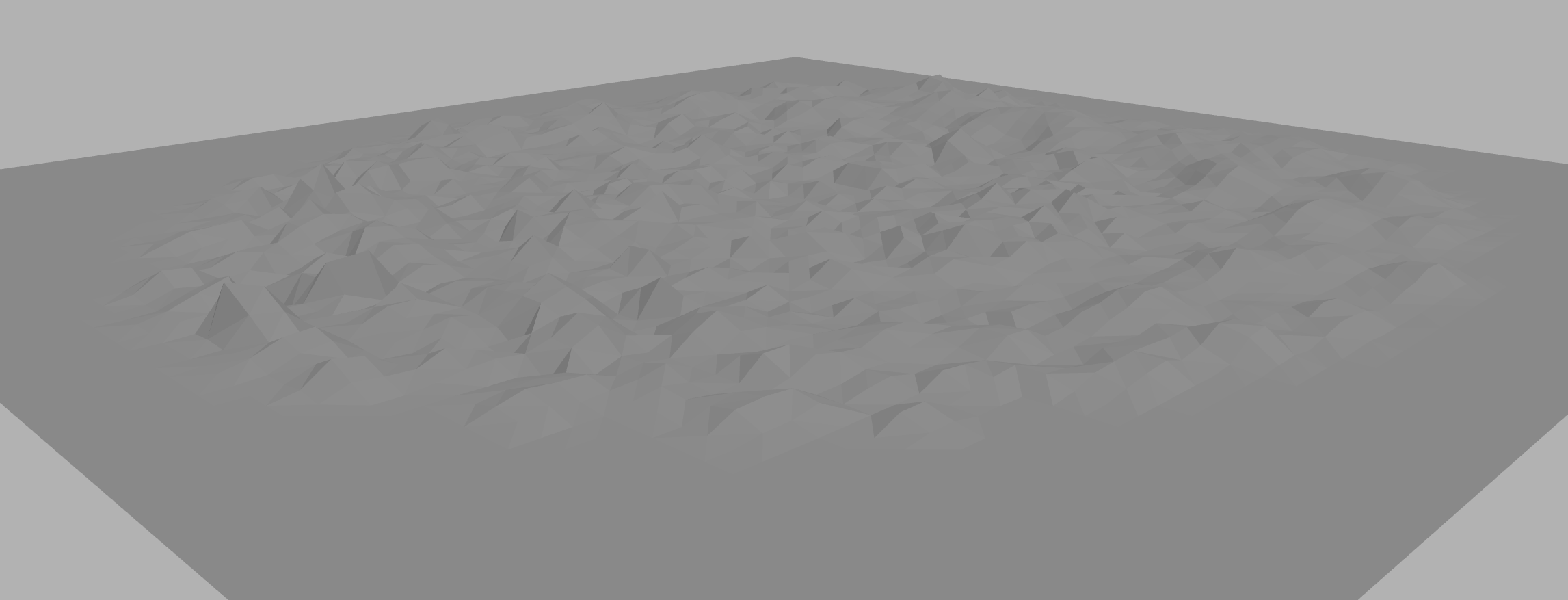}
    	\caption{
    	    Rough environment
        }
        \label{fig:rough-env}
    \end{subfigure}
    \caption{
        Environments used in experiments, simulated using Gazebo.
    }
    \label{fig:environments}
\end{figure}

In this work, we will use two different environments: a plain environment (Figure \ref{fig:plane-env}) and a rough environment (Figure \ref{fig:rough-env}).
The plain environment is a simple plane which is considered an easy environment to locomote in, while the rough environment consists of a polygon mesh with quadrilateral faces to represent small hills, which is considered a harder environment to locomote in due to its increased surface roughness.
Our first experiment is only conducted with the plain environment due to limitations in the simulation software regarding the interaction between the linear actuator module and the rough environment.
Furthermore, the second experiment is conducted without the linear actuator module for the same reason.

\subsection{Evolution}
For the evolution of the robots, the steady-state ($\mu + \lambda$) strategy is used \cite{eiben2003introduction}.
A population of $\mu = 100$ is evolved for 300 generations.
Each generation, $\lambda = 50$ offspring are created through tournament selection with two parents, with each set of parents producing one offspring.
From the pool of $\mu + \lambda$ individuals, $\mu$  individuals are selected for the next generation through binary tournament selection.
This process is repeated 20 times for each experiment.

\subsection{Fitness Function}
We consider the task of directed locomotion for all our experiments, which is considered a difficult and unsolved problem~\cite{sproewitz2008learning}.
The fitness function used for directed locomotion is taken from \cite{lan2018directed}:
\begin{equation} \label{eqn:fitness}
    f = \frac{\lvert distProjection \rvert}{lengthTraj + \epsilon} \left[\frac{distProjection}{\delta + 1} - penalty \right],
\end{equation}
where $distProjection$ is the distance traveled by the robot in the target direction, $lengthTraj$ is the length of the trajectory of the robot, $\epsilon$ is an infinitesimal constant, $\delta$ is the deviation between the robot's locomotion direction and the target direction, and $penalty$ is a term that penalizes deviating from the target direction.
For a complete description of these values, see \cite[Section 4]{lan2018directed}.
In our experiments, we set the target $30^\circ$ to the east relative to the robot ($\beta_0 = \pi/3$).
Robots are given an evaluation time of 30 seconds, after which Equation~\ref{eqn:fitness} is calculated.

\begin{table}[ht]
    \centering
    \caption{
        Experimental parameters of the \gls{ea} used.
    }
    \begin{tabular}{cc}
		\hline
		Representation & \gls{cppn} \\
		\hline
		Fitness function & Directed locomotion ($30^\circ$ to east) \\
		\hline
		Parent selection & Binary tournament selection \\
		\hline
		Survival selection & Steady state ($\mu + \lambda$) \\
		\hline
		Population size & 100 \\
		\hline
		Number of offspring & 50 \\
		\hline
		Initialization & Random \\
		\hline
		Termination condition & 300 generations \\
		\hline
		Evaluation time & 30 seconds \\
		\hline
		Runs & 20 \\
		\hline
	\end{tabular}
    \label{tab:experimental-setup}
\end{table}

\section{Experimental Results and Discussion}
This section discusses the results of the two aforementioned experiments.\footnote{Some examples of well-performing evolved robots are illustrated at \url{https://youtu.be/jF64aiIBqac}}
First, we investigate the effects of adding the linear actuator module in Experiment 1.
Second, we look at how the morphologies evolved in a plain environment differ from morphologies evolved in a rough environment in Experiment 2.
We will analyze the results by comparing the mean of the morphological descriptors of the last generation using the Wilcoxon signed-rank test~\cite{conover1999practical}.

\subsection{Experiment 1: Linear Actuator vs. No Linear Actuator}
\begin{figure}[h]
    \centering
    \begin{subfigure}{.485\linewidth}
      \includegraphics[width=1\linewidth]{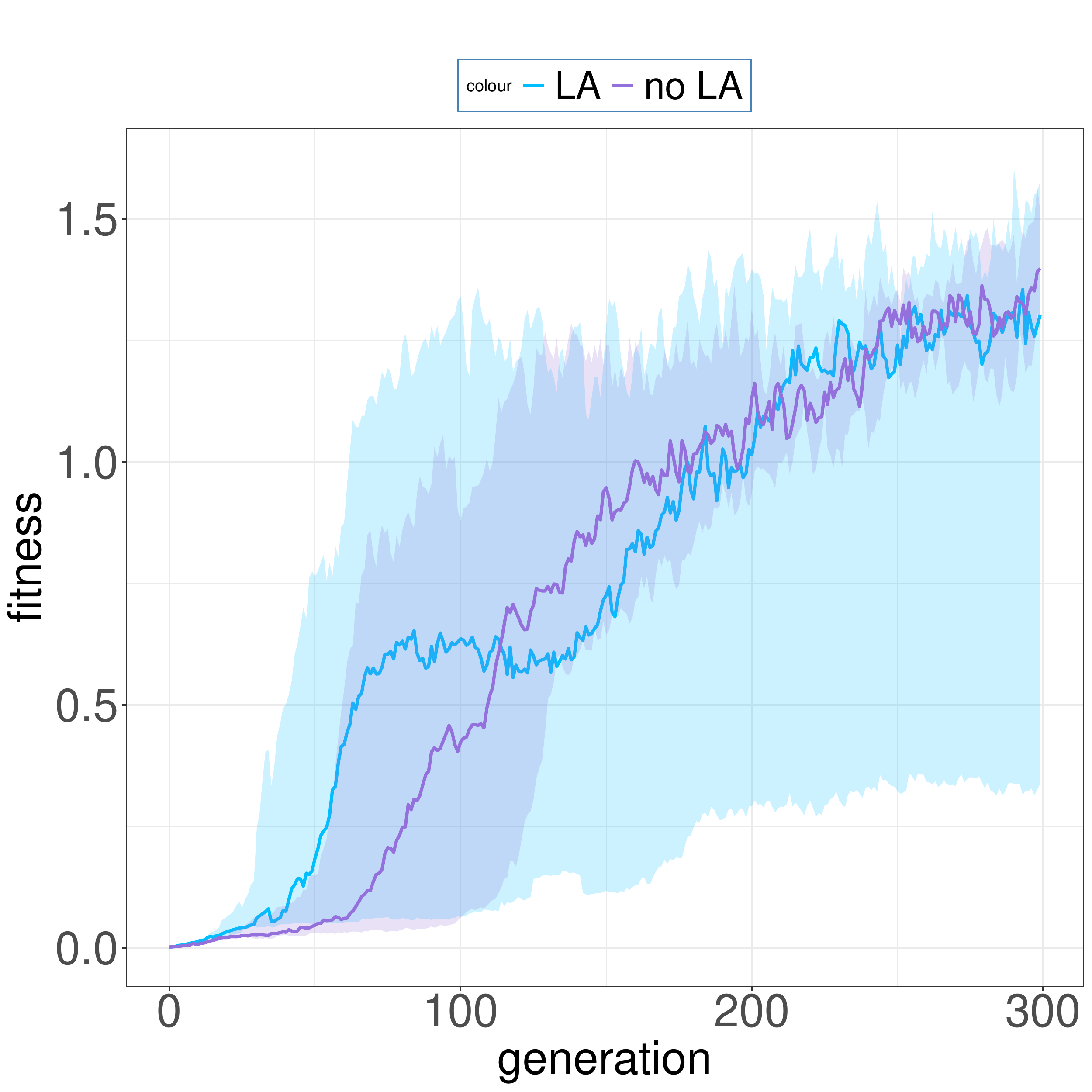}
      \caption{
        Mean fitness
      }
      \label{fig:la-nola-mean-fitness}
    \label{fig:over-generations}
    \end{subfigure}\hfill
    \begin{subfigure}{.485\linewidth}
        \includegraphics[width=1\linewidth]{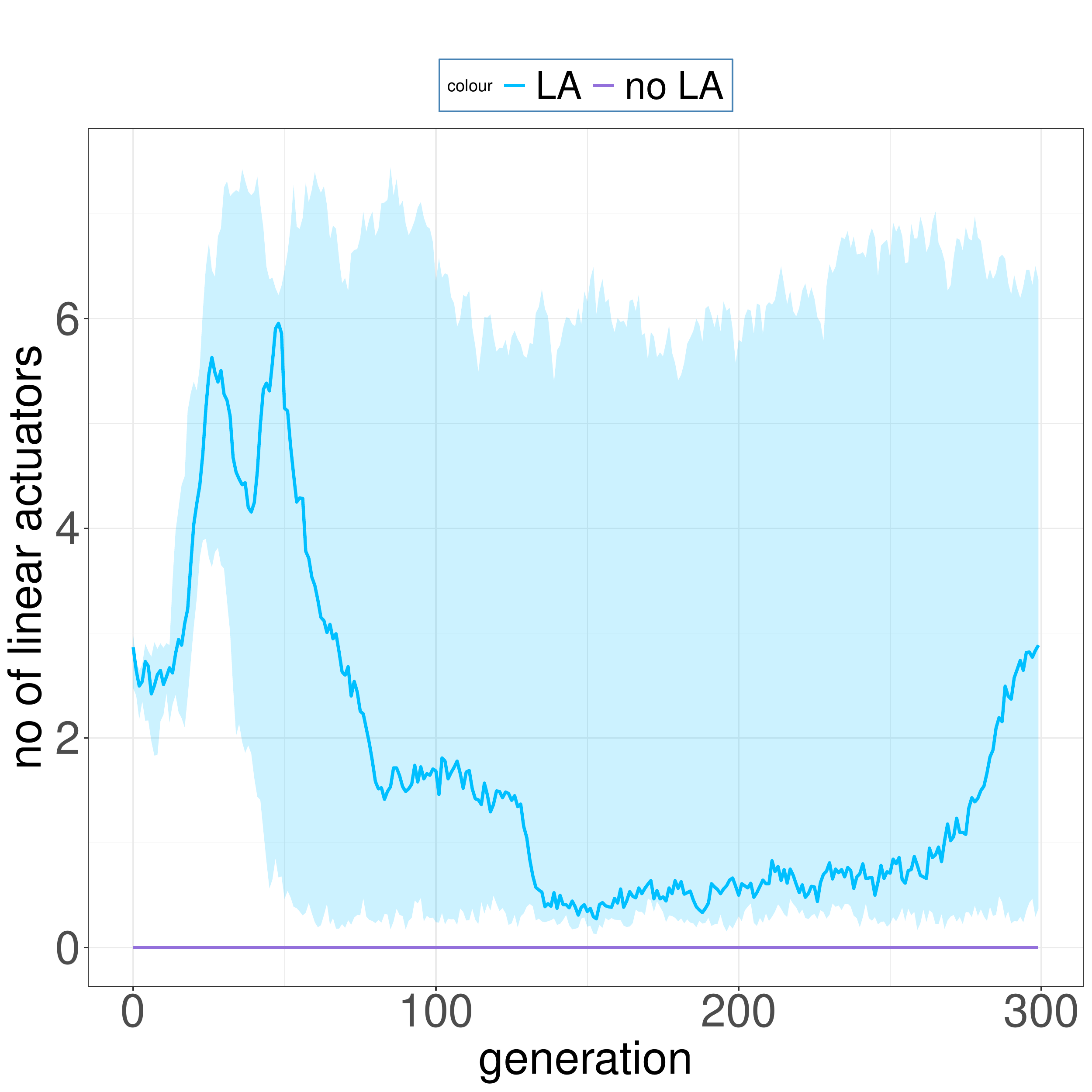}
        \caption{
          Mean no. linear actuators
        }
        \label{fig:la-nola-la-count}
    \end{subfigure}
    \caption{
        Progression of the mean fitness and number of linear actuators over all runs.
        Envelopes around the curves represent the lower and upper quartiles.
    }
    \label{fig:over-generations}
\end{figure}

For the first experiment, we begin with looking at the mean of the morphological descriptors of the last generation of both populations.
Table \ref{tab:wilcoxon-la-vs-nola} shows the difference in morphological descriptors between robots evolved with and without the linear actuator module in the plain environment.
Furthermore, we plot the progression of the fitness and number of linear actuator modules of both populations in Figure \ref{fig:la-nola-mean-fitness} and \ref{fig:la-nola-la-count} respectively.

We find no significant differences in the morphological descriptors of robots evolved with the linear actuator module and without the linear actuator module (all $p > 0.05$).
Furthermore, there is no significant difference in mean fitness of the final generation ($p = 0.85$).
We thus conclude that adding the linear actuator module has no significant impact on the performance and morphologies of robots evolved for directed locomotion in a plain environment.

\begin{table}[h]
	\caption{
	    Wilcoxon tests $p$-values for the difference in the mean of every morphological descriptor in the final generation of the experiments with and without the linear actuator module in the plain environment.
	}
    \begin{tabular}{cc}
        \hline
        & LA vs. No LA \\
    	\hline
    	& $p <$ \\
    	\hline
    	Branching & 0.82 \\
    	\hline
    	Coverage & 0.12 \\
    	\hline
    	Relative number of joints & 0.063 \\
    	\hline
    	Relative number of limbs & 0.31 \\
    	\hline
    	Relative length of limbs & 0.21 \\
    	\hline
    	Proportion & 0.35 \\
    	\hline
    	Absolute size & 0.85 \\
    	\hline
    	Symmetry & 0.85 \\
    	\hline
	\end{tabular}
	\label{tab:wilcoxon-la-vs-nola}
\end{table}

\begin{figure}[ht]
\centering
\begin{subfigure}{.5\linewidth}
  \centering
  \includegraphics[width=1\linewidth]{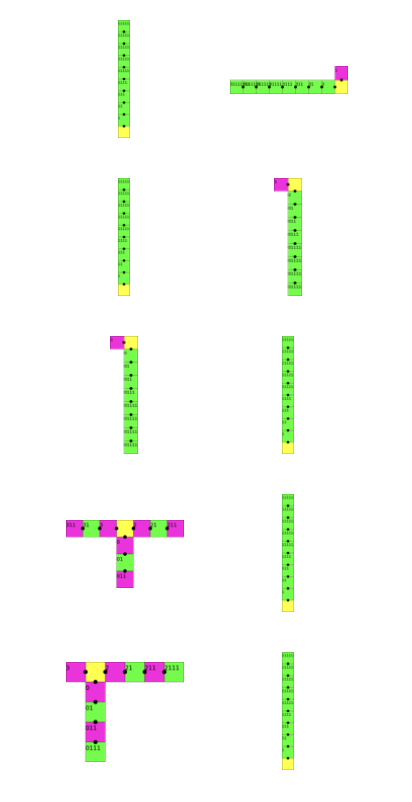}
  \caption{No \gls{la}}
  \label{fig:morph-nola}
\end{subfigure}%
\begin{subfigure}{.5\linewidth}
  \centering
  \includegraphics[width=1\linewidth]{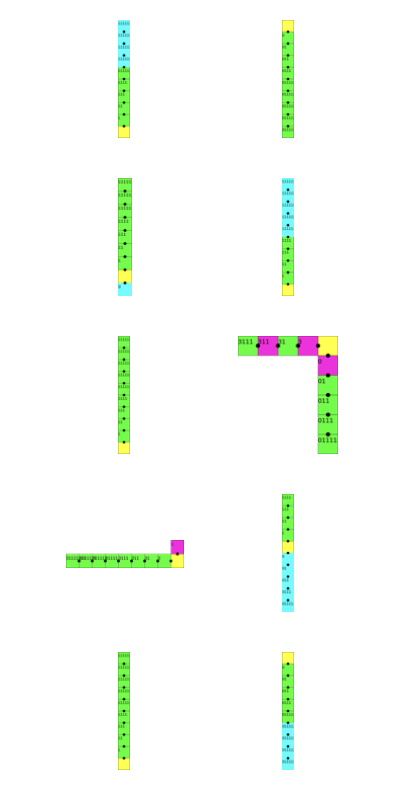}
  \caption{\gls{la}}
  \label{fig:morph-la}
\end{subfigure}
\caption{Best ten morphologies of the final generations of 20 runs evolved in the plain environment.}
\label{fig:morph-exp1}
\end{figure}

However, we find that the linear actuator remains in the final generation (Figure \ref{fig:la-nola-la-count}), though with no significant benefit.
This is further visible in Figure \ref{fig:morph-exp1}, where we visualize the morphologies of the best individuals from both populations.
The majority of the resulting morphologies for this experiment are `snake'-like as observed in previous works \cite{10.3389/frobt.2021.672379,pool2021moonwalkers,miras2019effects}.
In these `snake'-like morphologies, swapping some active hinges for linear actuators does not change its behavior as is further visible in the video, explaining why the linear actuator still gets selected but does not affect the fitness.
Potentially, the linear actuator is not distinct enough from the active hinge in its current form.
Further experiments with the length and weight of the linear actuator is needed to fully explore the potential of the linear actuator module.

\subsection{Experiment 2: Plain vs. Rough Environment}
\begin{figure*}[ht]
    \centering
    \begin{subfigure}{.18\textwidth}
      \centering
      \includegraphics[width=.85\linewidth]{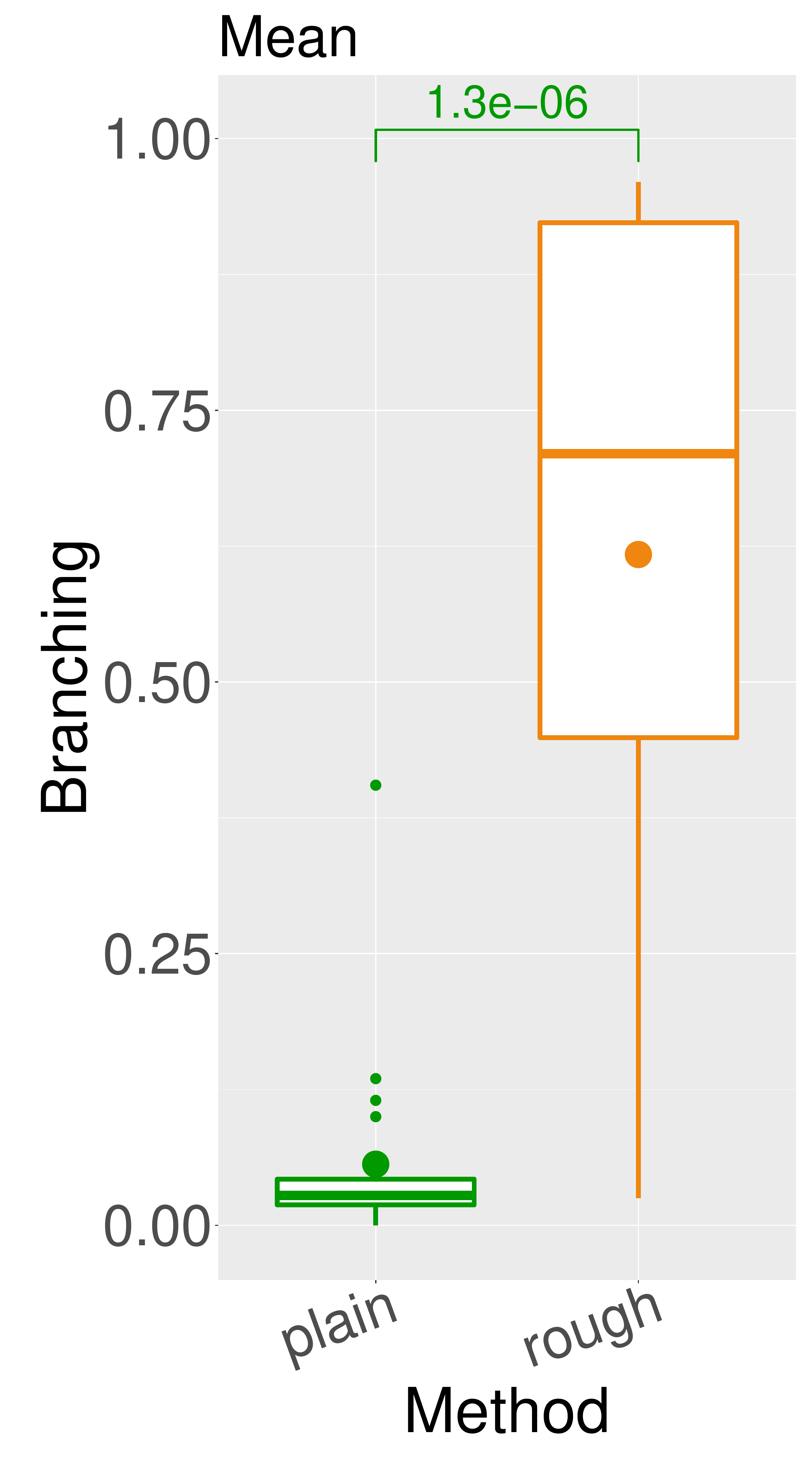}
    \end{subfigure}%
    \begin{subfigure}{.18\textwidth}
      \centering
      \includegraphics[width=.85\linewidth]{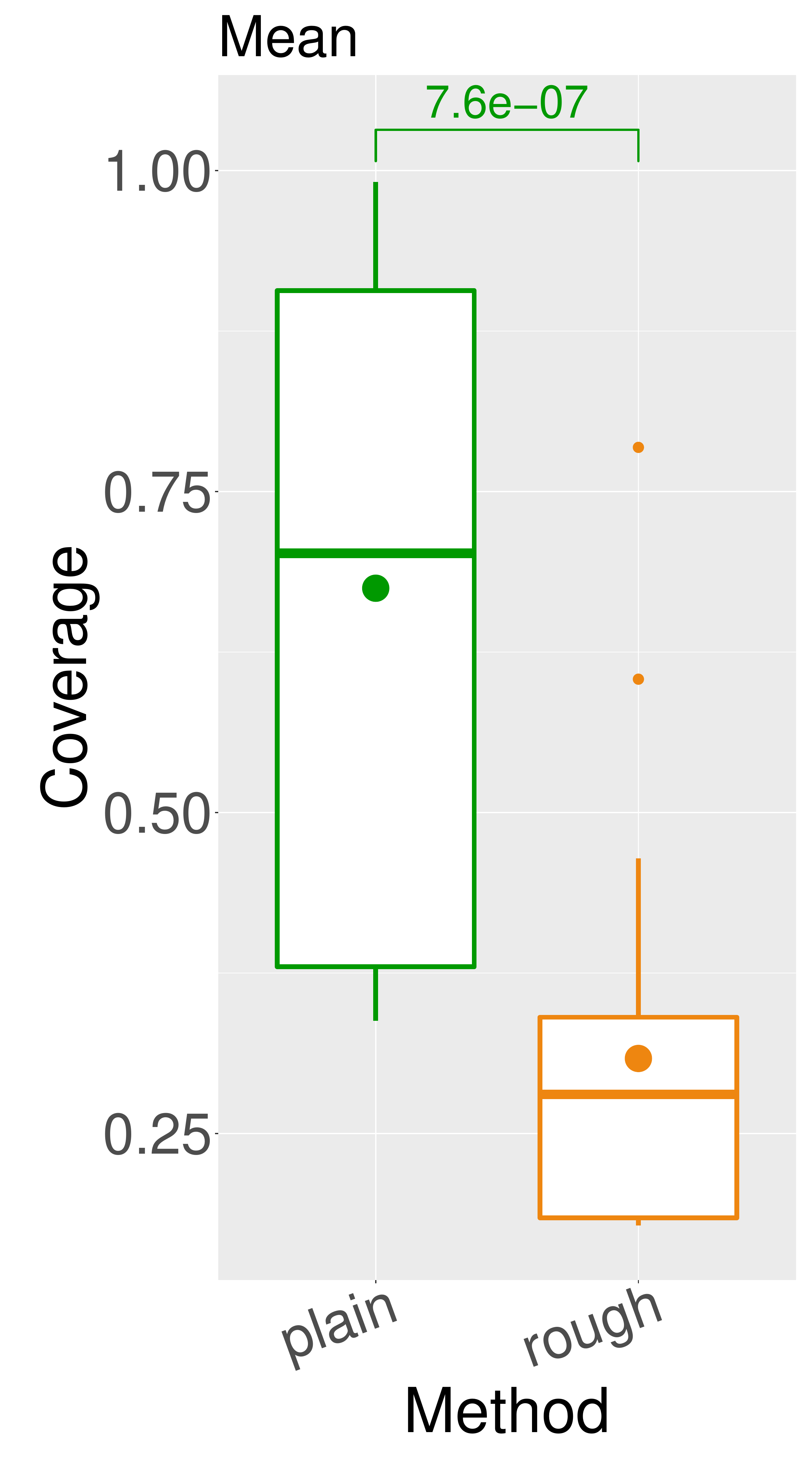}
    \end{subfigure}%
    \begin{subfigure}{.18\textwidth}
      \centering
      \includegraphics[width=.85\linewidth]{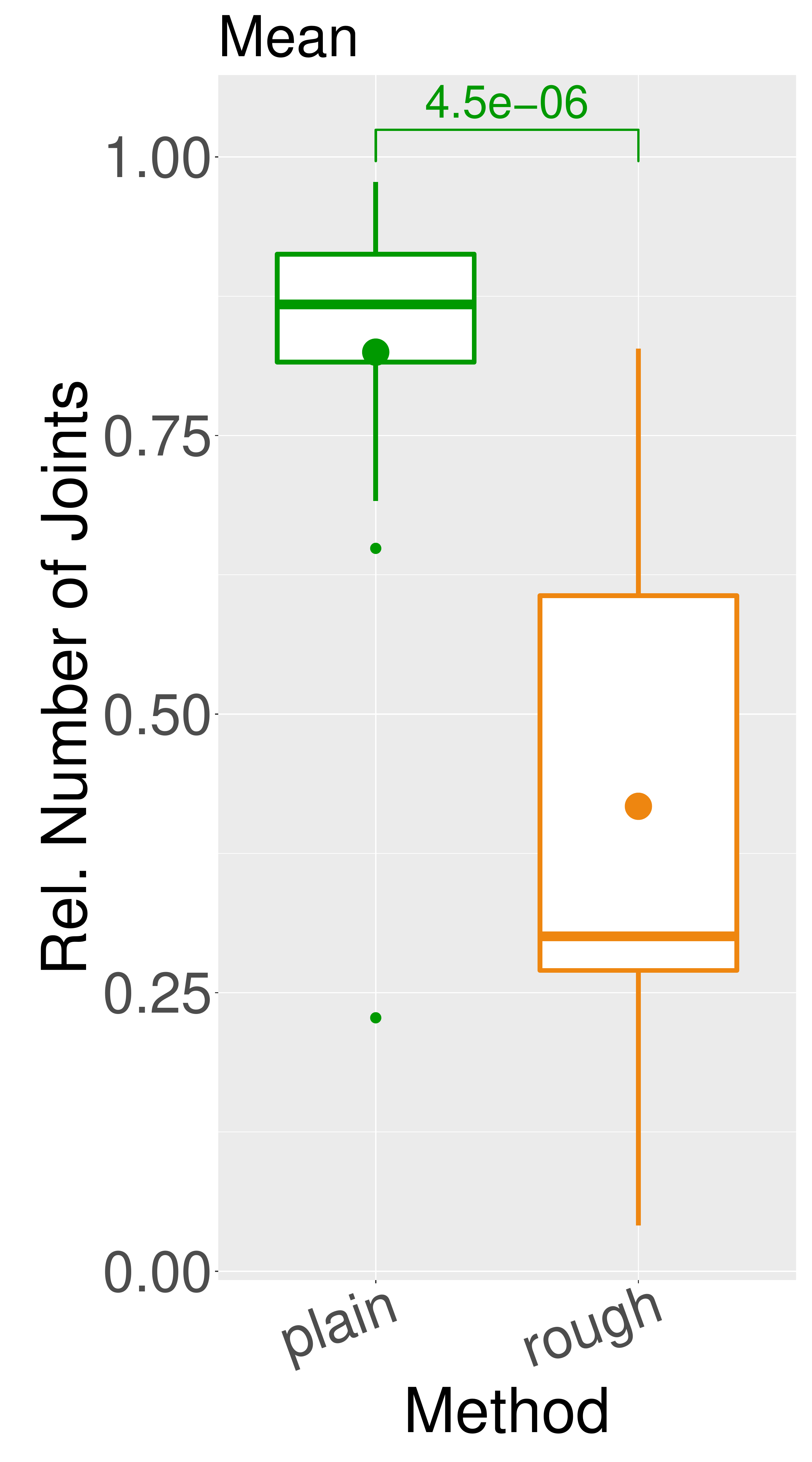}
    \end{subfigure}%
    \begin{subfigure}{.18\textwidth}
      \centering
      \includegraphics[width=.85\linewidth]{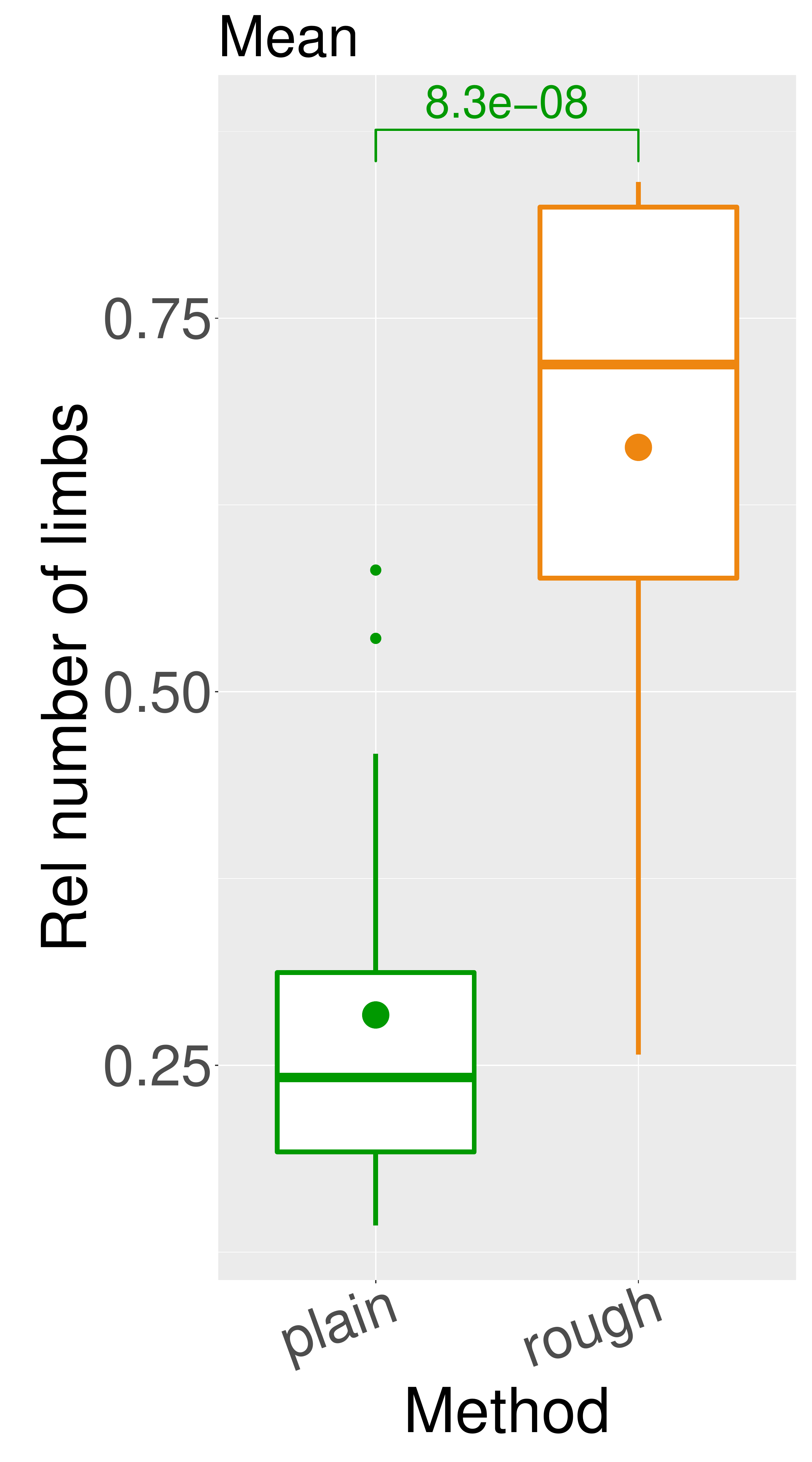}
    \end{subfigure}%
    \\
    \begin{subfigure}{.18\textwidth}
      \centering
      \includegraphics[width=.85\linewidth]{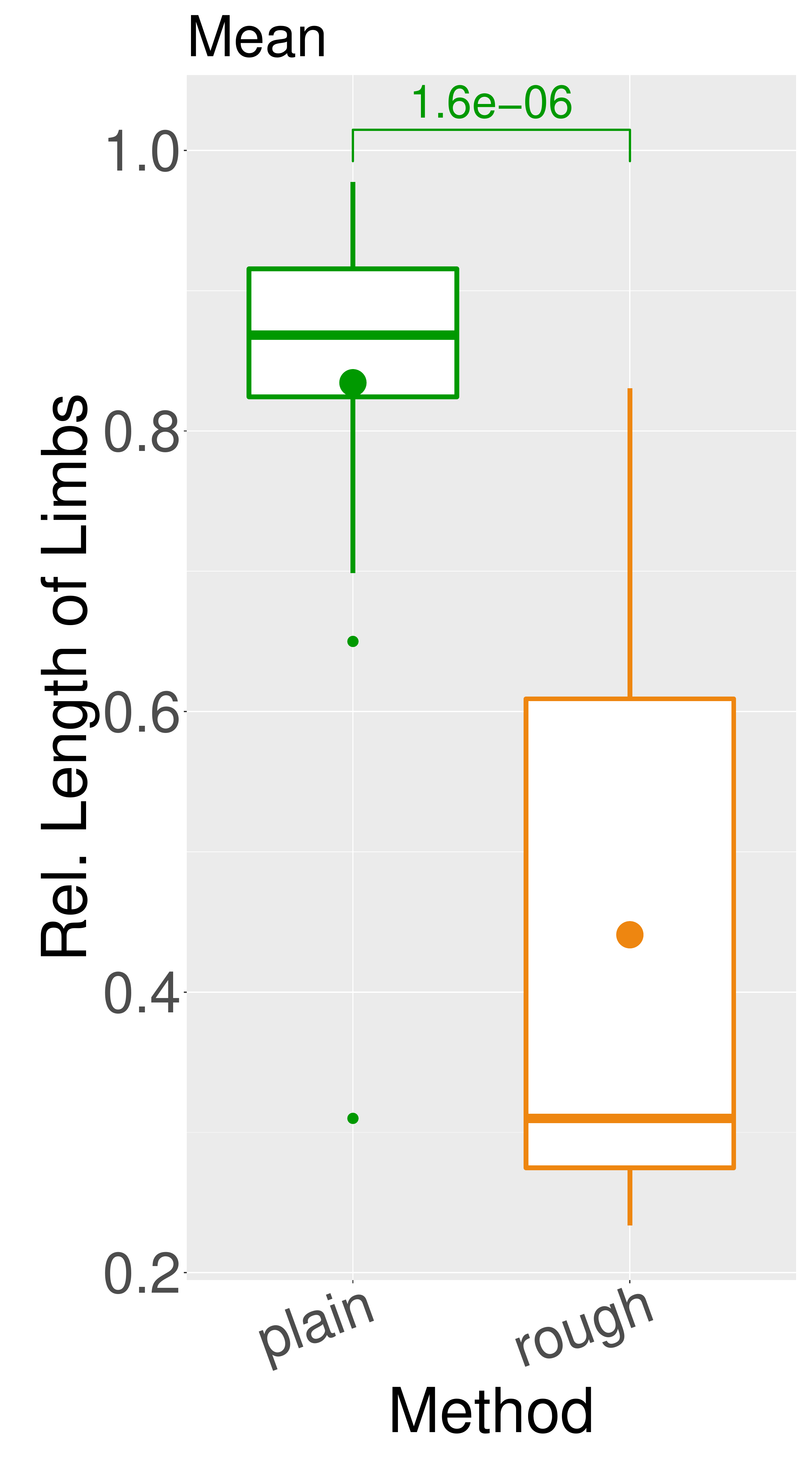}
    \end{subfigure}%
    \begin{subfigure}{.18\textwidth}
      \centering
      \includegraphics[width=.85\linewidth]{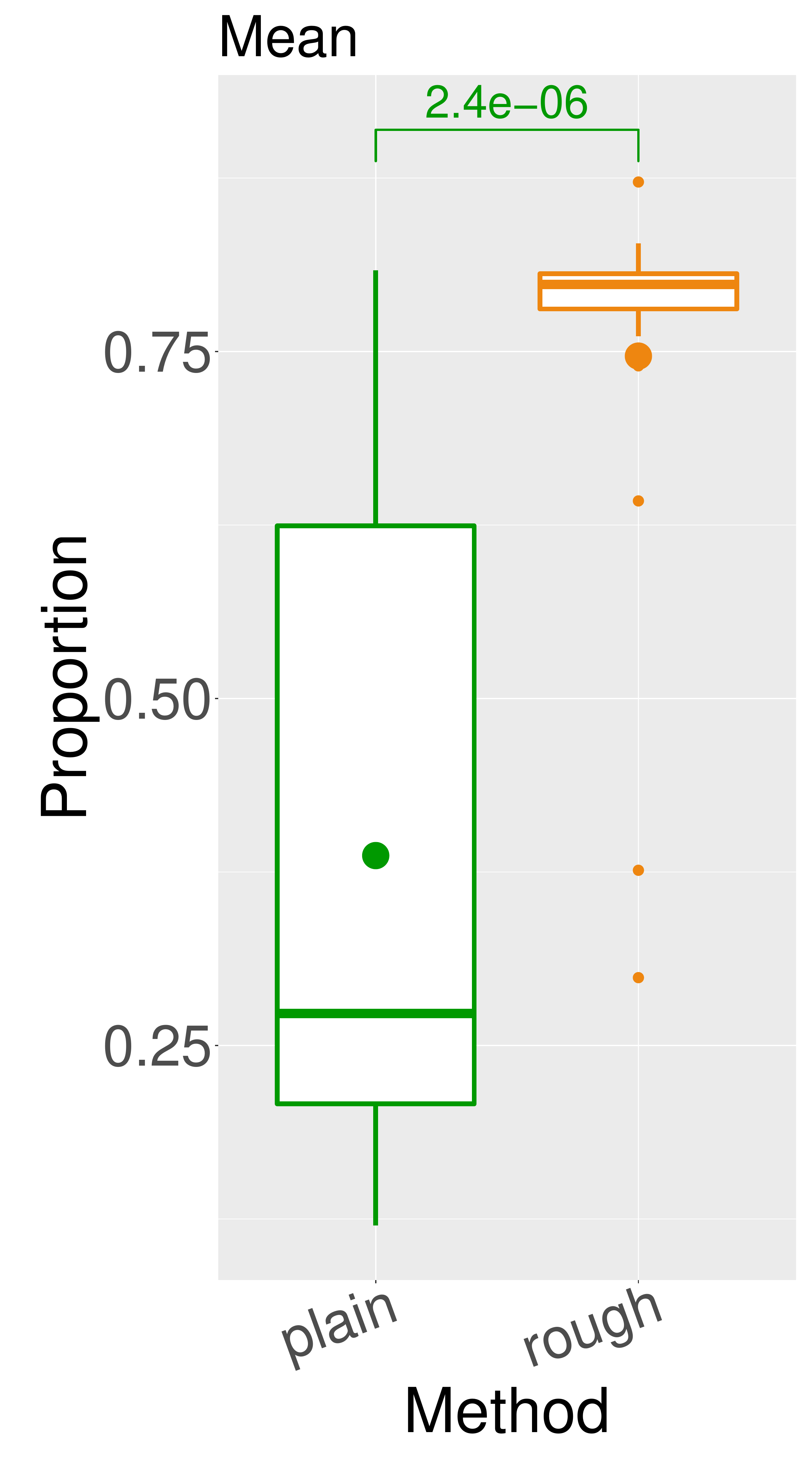}
    \end{subfigure}%
    \begin{subfigure}{.18\textwidth}
      \centering
      \includegraphics[width=.85\linewidth]{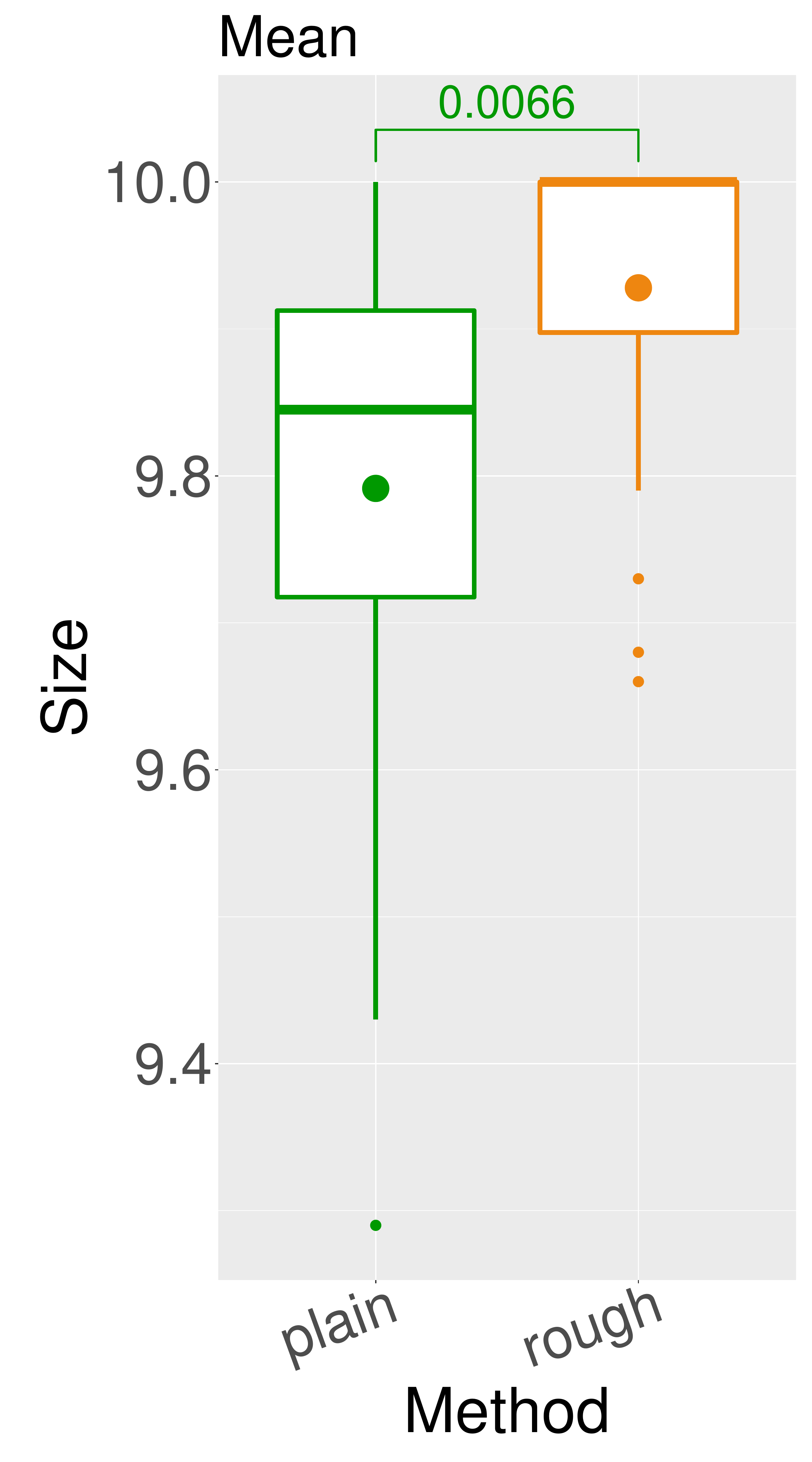}
    \end{subfigure}%
    \begin{subfigure}{.18\textwidth}
      \centering
      \includegraphics[width=.85\linewidth]{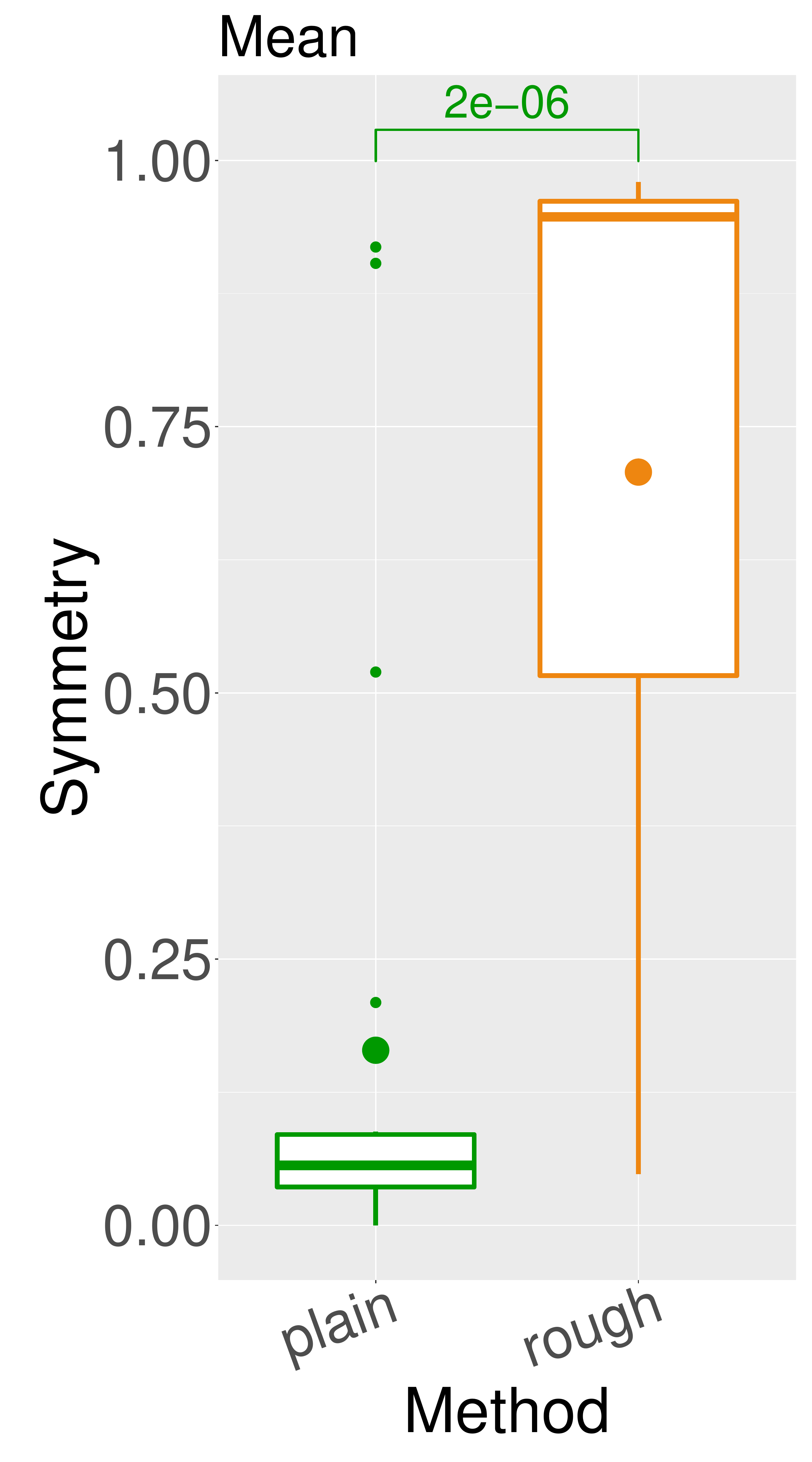}
    \end{subfigure}%
    \caption{Box plots comparing the mean of the morphological descriptors in the last generation of robots evolved in the plain and rough environment.}
    \label{fig:exp2-boxplots}
\end{figure*}

In the second experiment, we compare the morphologies of robots evolved in the plain environment against the morphologies of robots evolved in the rough environment.
The linear actuator module is omitted from this experiment due to limitations in the simulation software.
We compare the mean of the morphological descriptors of the last generation of both populations in Table \ref{tab:exp2}.
The distribution of the results is further visualized as box plots in Figure \ref{fig:exp2-boxplots}.
We find significant differences in all eight morphological descriptors between the morphologies of robots evolved in the plain and rough environment.
That is, significantly different morphologies have evolved in the rough environment compared to the plain environment for the same task of directed locomotion.

\begin{figure}[ht]
\centering
  \includegraphics[width=0.5\linewidth]{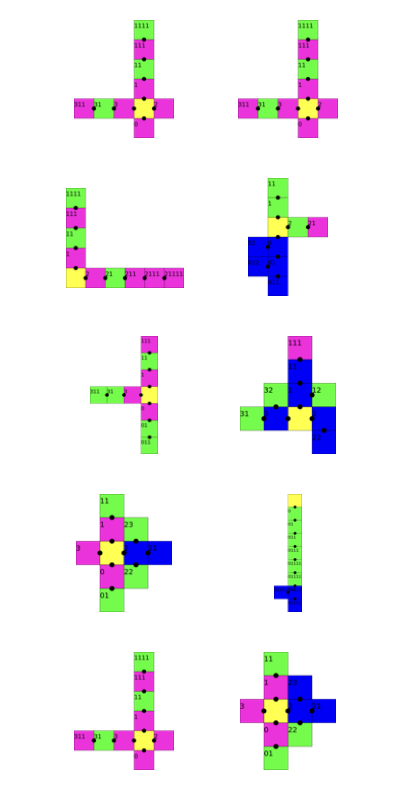}
  \caption{Best ten morphologies of the final generations of 20 runs evolved in the rough environment.}
  \label{fig:morph_rough}
\end{figure}

The morphologies evolved in the rough environment (Figure \ref{fig:morph_rough}) appear more diverse and complex than morphologies evolved in the plain environment (Figure \ref{fig:morph-nola}).
Furthermore, the morphologies evolved in the rough environment show more complex behavior.
It is visible from the video demonstration that they tend to use their limbs to grab hold of hills in the rough terrain to move, showing more interesting locomotion patterns.

From this experiment, we can conclude that the environment has a significant impact on the emergence of more complex morphologies and behaviors.
In a simple environment, simple morphologies often turn out to be sufficient to succeed.
As we have demonstrated in this work, more complex morphologies and behavior can emerge when changing one environmental variable: the terrain.
In turn, we present an in silico demonstration of environmental adaptation.

\begin{table}[ht]
    \caption{
	    Wilcoxon tests $p$-values for the difference in the mean of every morphological descriptor in the final generation of the experiments in the plain and rough environment without the linear actuator module.
	    Significant differences ($p < 0.05$) are boldfaced.
	}
    \begin{tabular}{cc}
        \hline
        & Plain vs. Rough \\
    	\hline
    	& $p <$ \\
    	\hline
    	Branching & $\mathbf{1.3 \cdot 10^{-6}}$ \\
    	\hline
    	Coverage & $\mathbf{7.6 \cdot 10^{-7}}$ \\
    	\hline
    	Relative number of joints & $\mathbf{4.5 \cdot 10^{-6}}$ \\
    	\hline
    	Relative number of limbs & $\mathbf{8.3 \cdot 10^{-8}}$ \\
    	\hline
    	Relative length of limbs & $\mathbf{1.6 \cdot 10^{-6}}$ \\
    	\hline
    	Proportion & $\mathbf{2.4 \cdot 10^{-6}}$ \\
    	\hline
    	Absolute size & $\mathbf{0.0066}$ \\
    	\hline
    	Symmetry & $\mathbf{2 \cdot 10^{-6}}$ \\
    	\hline
	\end{tabular}
	\label{tab:exp2}
\end{table}

\section{Conclusion and Future Work}
\Acrlong{ec} is an important theory in the field of \gls{er} which states that intelligence is the result of the interaction between the body (morphology), brain (controllers), and environment. In ER, most research focuses mainly on the evolution of robot controllers. In this research, attention is paid to the other two elements of the theory: the morphology and the environment. First, the number of modules of which robot morphologies can consist is expanded with the introduction of the linear actuator. Here, it is investigated how the linear actuator influences the performance and the morphology of the robots. Second, it is investigated how the environment influences the robot’s morphology. This is done by comparing the morphologies of robots evolved in two distinct environments, namely in an environment with a flat surface and one with a rough surface. For both experiments, changes in the body are assessed using eight morphological descriptors.

The results of our first experiment show that there is no significant difference for any of the eight morphological descriptors of robots evolved with or without the linear actuator. ‘Snake’-like morphologies are dominant in both cases. Besides, there is no significant difference in fitness. Therefore, we can answer the first research question, namely that adding the linear actuator to the morphological modules does not change the morphology of the robots and does not improve their performance. The first result could be explained by the observation made by Auerbach et al. \cite{auerbach2012relationship2}, namely that the flat terrain is too simplistic and that a simple morphology is enough to be successful in the task environment. For the latter result, this suggests that the linear actuator offers no advantages. However, when looking at the results of the second experiment, we see that the morphologies of robots evolved in the rough environment differ significantly for all morphological descriptors to those evolved in the plain environment. This allows us to answer our second research question, namely that robots evolved in the rough environment are more complex and diverse in both morphology and behavior. This is in line with one of the main principles of evolution, namely that evolution in different environments leads to different morphologies.

Although the results of this study suggest that adding the linear actuator had no benefit, we would not dismiss it just yet. For future work, we recommend evolving robots with the linear actuator in the rough environment. This allows you to investigate whether the linear actuator offers an advantage in other environments. In addition, we recommend investigating different lengths of the linear actuator, for example, one of 10 cm. It can be concluded from the results of Experiment 1, that the performance of the robots with a linear actuator is no better than the robots without the linear actuator. One possibility of this may be that the linear actuator does not bring significant advantages compared to the other four  modules. By investigating other lengths, it can be investigated whether the linear actuator does distinguish itself. In addition, it is advised to quantify the surface roughness and evolve in environments with different levels of roughness. This makes it possible to investigate how the surface roughness exactly influences the morphologies and allows to investigate whether there is a pattern. Furthermore, we think it will increase the practical relevance when testing with various fitness functions, such as climbing or obstacle avoidance. Finally, for the same reason, we recommend testing the effects of even more environmental properties, such as wind or the degree of friction on the surface.

\bibliographystyle{ACM-Reference-Format}
\bibliography{sample-base}

\end{document}